\documentclass[conference,a4paper]{APSIPA2012}

\usepackage{graphicx}
\usepackage{amssymb,amsmath,bm}
\usepackage{textcomp}
\usepackage{epsfig}
\usepackage{epstopdf, color}

\title{Multi-task Recurrent Model for Speech and Speaker Recognition}

\author{%
\authorblockN{%
Zhiyuan Tang\authorrefmark{2}\authorrefmark{3},
Lantian Li\authorrefmark{2} and
Dong Wang\authorrefmark{2}\authorrefmark{1}
}
\authorblockA{%
\authorrefmark{2}
Center for Speech and Language Technologies, Division of Technical Innovation and Development, \\
Tsinghua National Laboratory for Information Science and Technology\\
Center for Speech and Language Technologies, Research Institute of Information Technology, Tsinghua University
}
\authorblockA{%
\authorrefmark{3}
Chengdu Institute of Computer Applications, Chinese Academy of Sciences\\
E-mail: \{tangzy, lilt\}@cslt.riit.tsinghua.edu.cn\\
$^*$Corresponding author: wangdong99@mails.tsinghua.edu.cn}%
}

\begin{document}

  \maketitle
  \begin{abstract}

  Although highly correlated, speech and speaker recognition have been regarded as
  two independent tasks and studied by two communities. This is certainly not the way
  that people behave: we decipher both speech content and speaker traits at the same time.

  This paper presents a unified model to perform speech and speaker recognition
  simultaneously and altogether.
  The model is based on a unified neural network where the output of
  one task is fed to the input of the other, leading
  to a multi-task recurrent network. Experiments show that the joint model outperforms
  the task-specific models on both the two tasks.

  \end{abstract}

  \section{Introduction}

  Speech recognition (ASR) and speaker recognition (SRE) are two important research areas in speech processing. Traditionally, these
  two tasks are treated independently and studied by two independent communities, although some
  researchers indeed work on both areas. Unfortunately, this is not the way that human processes speech signals: we
  always decipher speech content and other meta information together and simultaneously, including languages, speaker
  characteristics, emotions, etc. This `multi-task decoding' is based on two foundations: (1) all these human capabilities
  share the same signal processing pipeline in our aural system, and (2) they are mutually beneficial
  as the success of one task promotes others' in real life. Therefore, we believe that multiple tasks in speech processing should
  be performed by a unified artificial intelligence system. This paper
  focuses on speech and speaker recognition, and demonstrates that these two tasks can be solved by a single unified model.

  In fact, the relevance of speech and speaker recognition has been recognized by researchers
  for a long time.
  On one hand, these two tasks share many common techniques, from the MFCC feature extraction
  to the HMM modeling; and on the other hand, researchers in both areas have been used to learning
  from each other. For instance, the success of deep neural networks (DNNs)
  in speech recognition~\cite{dahl2012context,hinton2012deep} has motivated the neural model in speaker
  recognition~\cite{variani2014deep,li2015improved}. Additionally,
  researchers also know for a long time that employing the knowledge provided by one area often helps improve
  the other. For instance, i-vectors produced by speaker recognition have been used to improve speech
  recognition~\cite{senior2014improving}, and phone posteriors
  derived from speech recognition have been utilized to improve speaker recognition~\cite{lei2014novel,kenny2014deep}.
  Moreover, the combination of these two systems has already gained attention.
  For instance, speech and speaker joint inference was proposed in~\cite{benzeghiba2003combination},
  and an LSTM-based multi-task model was proposed in~\cite{li2015modeling}.
  Although highly interesting, all the above research can not be considered as multi-task learning, and the speech
  and speaker recognition systems are designed, trained and executed independently.

  The development of deep learning techniques in speech processing provides new hope for multi-task learning.
  Since 2011, deep recurrent neural networks (RNNs) have become the new state-of-the-art architectures
  in speech recognition~\cite{graves2014towards,sak2014long}, and recently, the same architecture has
  gained much success in speaker recognition, at least in text-dependent conditions~\cite{heigold2015end}. In both
  the two tasks, deep learning
  delivers two main advantages: first, the structural depth (multiple layers) extracts task-oriented features, and second, the
  temporal depth (recurrent connections) accumulates dynamic evidence. Due to the similarity in the model
  structure, a simple question rises that can we use a single model to perform the two tasks together?

  Indeed, this `multi-task learning' has been known working well to boost correlated tasks~\cite{wang2015transfer}.
  For example, in
  multilingual speech recognition, it has been known that sharing low-level layers of DNNs can improve
  performance on each language~\cite{huang2013cross}. And in another experiment, phone and grapheme recognition were treated as two correlated tasks~\cite{chen2015multi}. The central idea
  of multi-task learning in the deep learning era is that correlated tasks can
  share the same feature extraction, and so the low-level layers of DNNs for
  these tasks can be shared. However, this feature-sharing architecture does not apply to
  speech and speaker recognition. This is because these two tasks
  are actually `negatively correlated': speech recognition requires features involving as much as content information,
  with speaker variance removed; while speaker recognition requires features involving as much as speaker information,
  with linguistic content removed. For these tasks, feature sharing is certainly not applicable.
  Unfortunately, many tasks are negatively correlated, e.g., language identification and speaker recognition, emotion recognition and speech recognition. Finding a multi-task learning approach that can deal with
  negatively-correlated tasks is therefore highly desirable.

\begin{figure}[t]
        \centering
        \includegraphics[width=0.8\linewidth]{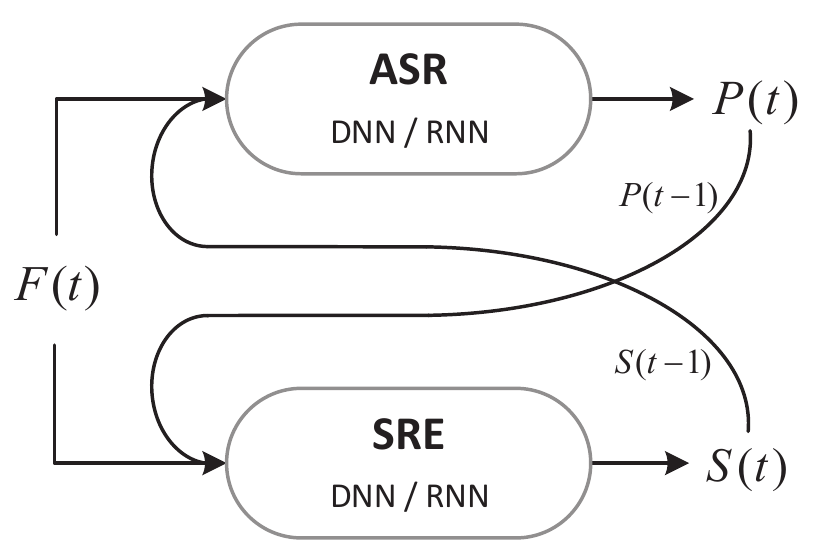}
        \caption{ Multi-task recurrent learning for ASR and SRE. $F(t)$ denotes primary features (e.g., Fbanks), $P(t)$ denotes phone identities (e.g., phone posteriors, high-level representations for phones), $S(t)$ denotes speaker identities (e.g., speaker posteriors, high-level representations for speakers).}
        \label{fig:diagram}
      \end{figure}

  This paper presents a novel recurrent architecture that can be used to learn negatively-correlated tasks simultaneously. The basic idea is to use the output of one task as part of the input of others. It would be ideal if the output of one task can provide information for others immediately,
  but this is not feasible in implementation. Therefore the output of one task at the previous
  time step is used to provide information for others at the current time step. This leads to an
  inter-task recurrent structure that is similar to conventional RNNs, though
  the recurrent connections link different tasks.
  We employed this multi-task recurrent learning to speech and speaker recognition
  and observed promising results. The idea is illustrated in Fig.~\ref{fig:diagram}.
  We note that a similar multi-task architecture
  was recently proposed in~\cite{li2015modeling}. The difference
  is that they focus on speaker adaptation for ASR,
  while we demonstrated improvement on both ASR and SRE
  tasks with the joint learning.

  The rest of the paper is organized as follows: Section~\ref{sec:model} presents the model architecture, and
  Section~\ref{sec:exp} reports the experiments. The conclusions plus the future work are presented in Section~\ref{sec:con}.

  \section{Models}
  \label{sec:model}

  \subsection{Basic single-task model}

\begin{figure}[ht]
        \centering
        \includegraphics[width=1\linewidth]{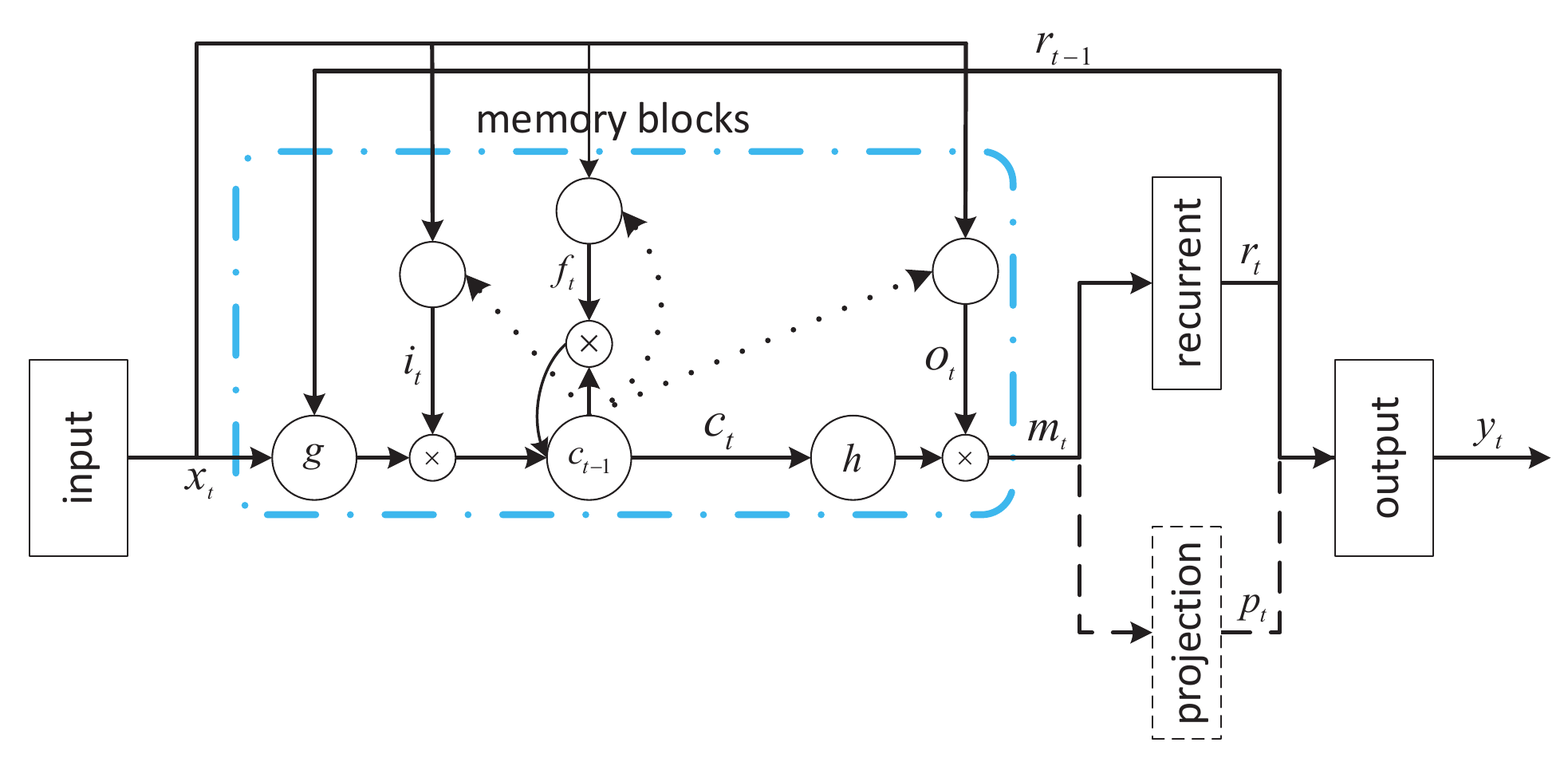}
        \caption{ Basic recurrent LSTM model for ASR and SRE single-task baselines.
        The picture is reproduced from~\cite{sak2014long}.}
        \label{fig:lstm}
  \end{figure}

  We start from the single-task models for ASR and SRE. As mentioned, the state-of-the-art architecture
  for ASR is the recurrent neural network, especially the long short-term memory (LSTM)~\cite{graves2014towards}. This model also
  delivers good performance in SRE~\cite{heigold2015end}. We therefore choose LSTM to build the baseline
  single-task systems. Particularly, the modified LSTM structure proposed in~\cite{sak2014long} is used. The network structure is shown in Fig.~\ref{fig:lstm}.

  The associated computation is as follows:
  \begin{eqnarray}
    i_t &=& \sigma(W_{ix}x_{t} + W_{ir}r_{t-1} + W_{ic}c_{t-1} + b_i) \nonumber\\
    f_t &=& \sigma(W_{fx}x_{t} + W_{fr}r_{t-1} + W_{fc}c_{t-1} + b_f) \nonumber\\
    c_t &=& f_t \odot c_{t-1} + i_t \odot g(W_{cx}x_t + W_{cr}r_{t-1} + b_c) \nonumber\\
    o_t &=& \sigma(W_{ox}x_t + W_{or}r_{t-1} + W_{oc}c_t + b_o) \nonumber\\
    m_t &=& o_t \odot h(c_t) \nonumber\\
    r_t &=& W_{rm} m_t \nonumber\\
    p_t &=& W_{pm} m_t \nonumber\\
    y_t &=& W_{yr}r_t + W_{yp}p_t + b_y \nonumber
  \end{eqnarray}

  \noindent In the above equations, the $W$ terms denote weight matrices and those associated with cells were set to be diagonal in our
  implementation. The $b$ terms denote bias
  vectors. $x_t$ and $y_t$ are the input and output symbols respectively; $i_t$, $f_t$, $o_t$ represent respectively
  the input, forget and output gates; $c_t$ is the cell and $m_t$ is the cell output. $r_t$ and $p_t$ are two output components derived from $m_t$, where $r_t$ is recurrent and fed to the next time step, while $p_t$ is not recurrent and contributes to the present output only.
  $\sigma(\cdot)$ is the logistic sigmoid function, and $g(\cdot)$ and $h(\cdot)$ are non-linear activation functions, often chosen to be hyperbolic. $\odot$ denotes the element-wise multiplication.

  \subsection{Multi-task recurrent model}

\begin{figure}[ht]
        \centering
        \includegraphics[width=1\linewidth]{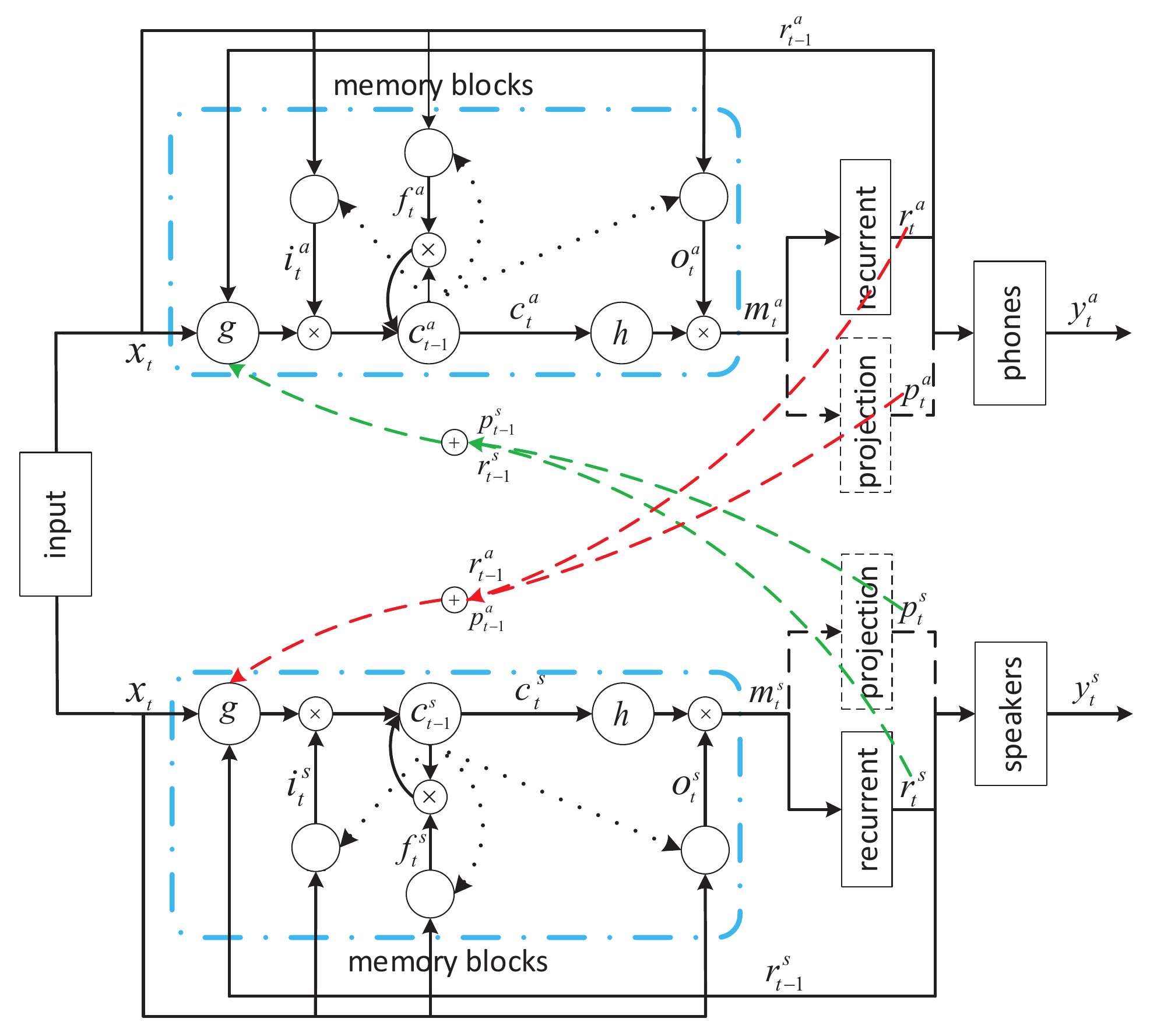}
        \caption{ Multi-task recurrent model for ASR and SRE, an example.}
        \label{fig:multi}
  \end{figure}

  The basic idea of the multi-task recurrent model, as shown in Fig.~\ref{fig:diagram}, is to use the output of
  one task at the current frame as an auxiliary information to supervise other tasks when processing the next frame.
  When this idea is materialized as a computational model, there are many alternatives that need to be carefully investigated. In this study, we use the recurrent LSTM model shown in the previous section to build the ASR
  component and the SRE component, as shown in Fig.~\ref{fig:multi}. These two components are identical
  in structure and accept the same input signal. The only difference is
  that they are trained with different targets, one for phone discrimination and the other for speaker discrimination.
  Most importantly, there are some inter-task recurrent links that combine the two components as a single
  network, as shown by the dash lines in Fig.~\ref{fig:multi}.

  Besides the model structure, a bunch of design options need to be chosen. The first question is where the recurrent information should be extracted. For example, it can be extracted from the
  cell $c_t$ or cell output $m_t$, or from the output component $r_t$ or $p_t$, or even from the output $y_t$. Another
  question is which computation block will receive the recurrent information. It can be simply the input variable $x_{t}$, but can also be the input gate $i_t$, the output gate $o_t$, the forget gate $f_t$ or the non-linear function
  $g(\cdot)$. Actually, augmenting the recurrent information to $x_{t}$ is equal to feed the information to $i_t$, $o_t$, $f_t$ and $g(\cdot)$ simultaneously. Note that a weight matrix is introduced as an extra free parameter
  for each recurrent information feedback.
  Moreover, the component that the information is extracted from is not necessarily the
  same for different tasks, nor is the component that receives the information. However in this study,
  we simply consider the symmetric structure.


  With all the above alternatives, the multi-task recurrent model is rather flexible.
  The structure shown in Fig.~\ref{fig:multi} is just one simple example, where the recurrent information
  is extracted from both the recurrent projection $r_t$ and the nonrecurrent projection $p_t$,
  and the information is applied to the non-linear function $g(\cdot)$. We use the
  superscript $a$ and $s$ to denote the ASR and SRE tasks respectively.
  The computation for ASR can be expressed as follows:

  \begin{eqnarray}
    i^a_t &=& \sigma(W^a_{ix}x_{t} + W^a_{ir}r^a_{t-1} + W^a_{ic}c^a_{t-1} + b^a_i) \nonumber\\
    f^a_t &=& \sigma(W^a_{fx}x_{t} + W^a_{fr}r^a_{t-1} + W^a_{fc}c^a_{t-1} + b^a_f) \nonumber\\
    g^a_t &=&  g(W^a_{cx}x^a_t + W^a_{cr}r^a_{t-1} + b^a_c +    \underline{ W^{as}_{cr}r^s_{t-1} + W^{as}_{cp}p^s_{t-1}  } ) \nonumber\\
    c^a_t &=& f^a_t \odot c^a_{t-1} + i^a_t \odot g^a_t \nonumber\\
    o^a_t &=& \sigma(W^a_{ox}x^a_t + W^a_{or}r^a_{t-1} + W^a_{oc}c^a_t + b^a_o) \nonumber\\
    m^a_t &=& o^a_t \odot h(c^a_t) \nonumber\\
    r^a_t &=& W^a_{rm} m^a_t \nonumber\\
    p^a_t &=& W^a_{pm} m^a_t \nonumber\\
    y^a_t &=& W^a_{yr}r^a_t + W^a_{yp}p^a_t + b^a_y \nonumber 
  \end{eqnarray}

  \noindent and the computation for SRE is as follows:
  \begin{eqnarray}
    i^s_t &=& \sigma(W^s_{ix}x_{t} + W^s_{ir}r^s_{t-1} + W^s_{ic}c^s_{t-1} + b^s_i) \nonumber\\
    f^s_t &=& \sigma(W^s_{fx}x_{t} + W^s_{fr}r^s_{t-1} + W^s_{fc}c^s_{t-1} + b^s_f) \nonumber\\
    g^s_t &=& g(W^s_{cx}x^s_t + W^s_{cr}r^s_{t-1} + b^s_c +    \underline{ W^{sa}_{cr}r^a_{t-1} + W^{sa}_{cp}p^a_{t-1}  } ) \nonumber\\
    c^s_t &=& f^s_t \odot c^s_{t-1} + i^s_t \odot g^s_t \nonumber\\
    o^s_t &=& \sigma(W^s_{ox}x^s_t + W^s_{or}r^s_{t-1} + W^s_{oc}c^s_t + b^s_o) \nonumber\\
    m^s_t &=& o^s_t \odot h(c^s_t) \nonumber\\
    r^s_t &=& W^s_{rm} m^s_t \nonumber\\
    p^s_t &=& W^s_{pm} m^s_t \nonumber\\
    y^s_t &=& W^s_{yr}r^s_t + W^s_{yp}p^s_t + b^s_y \nonumber 
  \end{eqnarray}

  \section{Experiments}
  \label{sec:exp}

  The proposed method was tested with the WSJ database, which has been labelled with both word transcripts and speaker identities. We first present the ASR and SRE baselines and then report the multi-task model. All the experiments were conducted with the Kaldi toolkit~\cite{povey2011kaldi}.

  \subsection{Data}

  \begin{itemize}
    \item Training set: This set involves $90\%$ of the speech data randomly selected from train\_si284
    (the other $10\%$ used for speaker identification test whose results for almost all systems were perfect thus not presented).
     It consists of $282$ speakers and $33,587$ utterances, with $40$-$144$ utterances per speaker.
     This set was used to  train the two LSTM-based single-task systems, an i-vector SRE baseline, and the proposed multi-task system.
  \end{itemize}

  \begin{itemize}
        \item Test set: This set involves three datasets (dev93, eval92 and eval93). It consists
        of $27$ speakers and $1,049$ utterances. This dataset was used to evaluate the performance of
        both ASR and SRE. For SRE, the evaluation consists of $21,350$ target trials and $528,326$
        non-target trials, constructed based on the test set.

  \end{itemize}

  \subsection{ASR baseline}

  The ASR system was built largely following the Kaldi WSJ s5 nnet3 recipe, except that we used a single LSTM layer for simplicity. The dimension of the cell was $1,024$, and the dimensions of the recurrent and nonrecurrent projections were set to $256$.  The target delay was $5$ frames. The natural stochastic gradient descent (NSGD) algorithm was employed to train the model~\cite{povey2014parallel}. The input feature was the $40$-dimensional Fbanks, with a symmetric $2$-frame window to splice neighboring frames. The output layer consisted of $3,419$ units, equal to the total number of pdfs in the conventional GMM system that was trained to bootstrap the LSTM model. The baseline performance is reported in Table~\ref{tab:asr-base}.

      \begin{table}[th]
        \caption{\label{tab:asr-base}  ASR baseline results.}
        \vspace{2mm}
        \centerline{
          \begin{tabular}{c|c|c|c|c}
            \hline
                   &dev92 & eval92 & eval93 & Total\\
            \hline
            WER\%  &8.36  & 5.14   & 8.06   & 7.41\\
            \hline
          \end{tabular}
        }
      \end{table}

  \subsection{SRE baseline}

  We built two SRE baseline systems: one is an i-vector system and the other is
  an `r-vector' system that is based on the recurrent LSTM model.

  For the i-vector system, the acoustic feature was $39$-dimensional MFCCs.
  The number of Gaussian components of the UBM was $1,024$, and the dimension of i-vectors was $200$.
  For the r-vector system, the architecture was similar to the one used by the LSTM-based ASR
  baseline, except that the dimension of the cell was $512$, and the dimensions
  of the recurrent and nonrecurrent projections were set to $128$.
  Additionally, there was no target delay. The input of the r-vector system was
  the same as ASR system, and the output was corresponding to the $282$ speakers in the
  training set.
  Similar to the work in~\cite{variani2014deep,li2015improved}, the speaker vector (`r-vector') was
  derived from the output of the recurrent and nonrecurrent projections,
  by averaging the output of all the frames. The dimension was $256$.

  The baseline performance is reported in Table~\ref{tab:sre-base}.
  It can be observed that the i-vector system generally outperforms the r-vector system.
  Particularly, the discriminative methods (LDA and PLDA) offer much
  more significant improvement for the i-vector system than
  for the r-vector system. This observation is consistent with the results reported
  in~\cite{li2015improved}, and can be attributed to the fact that
  the r-vector model has already been learned `discriminatively' with the LSTM structure.
  For this reason, we only consider the simple cosine kernel when scoring r-vectors
  in the following experiments.

      \begin{table}[th]
        \caption{\label{tab:sre-base} SRE baseline results.}
        \vspace{2mm}
        \centerline{
          \begin{tabular}{c|c|c|c}
            \hline
                          &\multicolumn{3}{|c}{EER\%} \\
            \hline
            System        &Cosine   & LDA    & PLDA \\
            \hline
            i-vector (200) &  2.89  &1.03    &0.57\\
            r-vector (256) &  1.84  &1.34    &3.18\\
            \hline
          \end{tabular}
        }
      \end{table}

  \subsection{Multi-task joint training}


  Due to the flexibility of the multi-task recurrent LSTM structure, it is not possible to
  evaluate all the configurations. We chose
  some typical ones and report the results in Table~\ref{tab:result}. We just show the ASR
  results on the combined dataset mentioned before. Note that the last configure, where
  the recurrent information is fed to all the gates and the non-linear activation
  $g(\cdot)$, is equal to
  augmenting the information to the input variable $x$.

    \begin{table}[thb!]
        \caption{\label{tab:result} Joint training results.}
        \vspace{2mm}
        \centerline{
          \begin{tabular}{|cc|cccc|c|c|}
            \hline
            \multicolumn{2}{|c|}{Feedback} & \multicolumn{4}{|c|}{Feedback}              & ASR     & SRE \\
            \multicolumn{2}{|c|}{Info. } & \multicolumn{4}{|c|}{Input}                   & WER\%     & EER\% \\
            \hline \hline
                         $r$ &  $p$       &  $i$   & $f$     &  $o$   &   $g$           &         &       \\
                    \hline
                             &            &        &         &        &                 &  7.41   &  1.84 \\
                    \hline
                    $\surd$  &            & $\surd$&         &        &                 &  7.05   &  0.62 \\
                    $\surd$  &$\surd$     & $\surd$&         &        &                 &  6.97   &  0.64 \\
                    \hline
                    $\surd$  &            &        &$\surd$  &        &                 &  7.12   &  0.66 \\
                    $\surd$  &$\surd$     &        &$\surd$  &        &                 &  7.24   &  0.65 \\
                    \hline
                    $\surd$  &            &        &         &$\surd$ &                 &  7.26   &  0.65 \\
                    $\surd$  &$\surd$     &        &         &$\surd$ &                 &  7.28   &  0.59 \\
                    \hline
                    $\surd$  &            &        &         &        & $\surd$         &  7.11   &  0.62 \\
                    $\surd$  &$\surd$     &        &         &        & $\surd$         &  7.11   &  0.67 \\
                    \hline
                    $\surd$  &            & $\surd$&$\surd$  &$\surd$ &                 &  7.06   &  0.66 \\
                    $\surd$  &$\surd$     & $\surd$&$\surd$  &$\surd$ &                 &  7.23   &  0.71 \\
                    \hline
                    $\surd$  &            & $\surd$&$\surd$  &$\surd$ &$\surd$          &  7.05   &  0.55 \\
                    $\surd$  &$\surd$     & $\surd$&$\surd$  &$\surd$ &$\surd$          &  7.23   &  0.62 \\
            \hline
          \end{tabular}
        }
      \end{table}

  From the results shown in Table~\ref{tab:result}, we first observe that
  the multi-task recurrent model consistently improves performance on both ASR and SRE,
  no matter where the recurrent information is extracted and where it applies.
  Most interestingly, on the SRE task, the multi-task system can obtain equal or
  even better performance than the i-vector/PLDA system.
  This is the first time that the two negatively-correlated tasks are learned jointly
  in a unified framework and boost each other.

  For the recurrent information, it looks like the recurrent projection $r_t$ is sufficient to
  provide valuable supervision for the partner task. Involving more information from
  the nonrecurrent projection does not offer consistent benefit. This observation, however, is only
  based on the present experiments. With more data, it is likely that more information leads
  to additional gains.

  For the recurrent information `receiver', i.e., the component that receives the recurrent information,
  it seems that for ASR the input gate and the activation function are equally effective, while
  the output gate seems not so appropriate. For SRE, all results seem good.
   Again, these observations are just based on a relative small database; with more
  data, the performance with different configurations may become distinguishable.


  \section{Conclusions}
  \label{sec:con}

  We report a novel multi-task recurrent learning architecture that can jointly train multiple
  negatively-correlated tasks. Primary results on the WSJ database demonstrated that the
  presented method can learn speech and speaker models simultaneously
  and improve the performance on both tasks. Future work involves analyzing more factors
  such as target delay, exploiting partially labelled data, and
  applying the approach to other negatively-correlated tasks.

\section*{Acknowledgment}
This work was supported by the National Science Foundation of China (NSFC)
Project No. 61371136, and the MESTDC PhD Foundation Project No.
20130002120011.


  \bibliographystyle{IEEEtran}
  \bibliography{joint}

\end{document}